\documentclass[runningheads]{llncs}
\usepackage{graphicx}
\usepackage{caption}
\usepackage{subcaption}
\usepackage{amsmath}
\usepackage{tabularx,booktabs}
\newcolumntype{Y}{>{\centering\arraybackslash}X}

\usepackage{ctable}
\usepackage{multirow}
\usepackage{acro}
\DeclareAcronym{VULD}{short=VLD, long=volumetric lesion detector}
\DeclareAcronym{VLD}{short=VLD, long=volumetric lesion detector}
\DeclareAcronym{PACS}{short=PACS, long=picture archiving and communication system}
\DeclareAcronym{P3D}{short=P3D, long=pseudo 3D}
\DeclareAcronym{P3DC}{short=P3DC, long=pseudo 3D convolution}
\DeclareAcronym{I3D}{short=I3D, long=inflated 3D}
\DeclareAcronym{ST3D}{short=ST-3D, long=spatio-temporal 3D}
\DeclareAcronym{ACS3D}{short=ACS-3D, long=axial-coronal-sagittal 3D}
\DeclareAcronym{IoU}{short=IoU, long=intersection over union}
\DeclareAcronym{AP}{short=AP, long=average precision} 
\DeclareAcronym{AR}{short=AR, long=average recall}
\DeclareAcronym{CT}{short=CT, long=computed tomography}
\DeclareAcronym{CADe}{short=CADe, long=computer-aided detection}
\DeclareAcronym{CADx}{short=CADx, long=computer-aided diagnosis}
\DeclareAcronym{PR}{short=DRP, long=deep representative points}
\DeclareAcronym{SPR}{short=SPR, long=surface point regression}
\DeclareAcronym{SOTA}{short=SOTA, long=state-of-the-art}
\DeclareAcronym{FP}{short=FP, long=false positive}
\DeclareAcronym{ULD}{short=ULD, long=universal lesion detection}
\DeclareAcronym{HCC}{short=HCC, long=hepatocellular carcinoma}
\DeclareAcronym{CNN}{short=CNN, long=convolutional neural network}
\DeclareAcronym{FROC}{short=FROC, long=free-response receiver operating characteristic}
\DeclareAcronym{TP}{short=TP, long=true positive} 
\DeclareAcronym{FPN}{short=FPN, long=feature pyramid network}
\def\Fig#1{{Fig.~\ref{fig:#1}}}

\def\Table#1{{Table~\ref{tbl:#1}}}
\def\eg{{e.g.}}
\def\etal{{et al.}}
\def\ie{{i.e.}}

\begin{document}
\title{Deep Volumetric Universal Lesion Detection using Light-Weight Pseudo 3D Convolution and Surface Point Regression}
\titlerunning{Deep VULD}
\author{Jinzheng Cai\inst{1} \and Ke Yan\inst{1} \and Chi-Tung Cheng\inst{2} \and Jing Xiao\inst{3} \and Chien-Hung Liao\inst{2} \and \\Le Lu\inst{1} \and Adam P. Harrison\inst{1}}
% index{Jinzheng, Cai}  
% index{Ke, Yan}  
% index{Chi-Tung, Cheng}  
% index{Jing, Xiao}  
% index{Chien-Hung, Liao}  
% index{Le, Lu}  
% index{Adam P., Harrison}  
%  
\authorrunning{J. Cai et al.}

\institute{PAII Inc., Bethesda, MD, USA \and
Chang Gung Memorial Hospital, Linkou, Taiwan, ROC \and 
Ping An Technology Co., Ltd., Shenzhen, China}

\maketitle              % typeset the header of the contribution

\begin{abstract}
Identifying, measuring and reporting lesions accurately and comprehensively from patient \acs{CT} scans are important yet time-consuming procedures for physicians. Computer-aided lesion/significant-findings detection techniques are at the core of medical imaging, which remain very challenging due to the tremendously large variability of lesion appearance, location and size distributions in 3D imaging. In this work, we propose a novel deep anchor-free one-stage \ac{VULD} framework that incorporates (1) \acl{P3DC} operators to recycle the architectural configurations and pre-trained weights from the off-the-shelf 2D networks, especially ones with large capacities to cope with data variance, and (2) a new \acl{SPR} method to effectively regress the 3D lesion spatial extents by pinpointing their representative key points on lesion surfaces. Experimental validations are first conducted on the public large-scale NIH DeepLesion dataset where our proposed method delivers new state-of-the-art quantitative performance. We also test \ac{VULD} on our in-house dataset for liver tumor detection. \ac{VULD} generalizes well in both large-scale and small-sized tumor datasets in CT imaging.


\end{abstract}
\acresetall
\section{Introduction}

Automated lesion detection is an important yet challenging task in medical image analysis, as exploited by \cite{Jiang2020ElixirNet,miccai/ShaoGMLZ19,Wang2020Towards,yan_2018_deeplesion,cvpr/yan18deeplesion,ijcai/ZhangXZCXS19,miccai/ZlochaDG19} on the public NIH DeepLesion dataset. Its aims include improving physician's reading efficiency and increasing the sensitivity for localizing/reporting small but vital tumors, which are more prone to be missed, \eg{} human-reader sensitivity is reported at $48\sim57\%$ with small-sized \ac{HCC} liver lesions~\cite{Addley_2011}. Automated lesion detection remains difficult due to the tremendously large appearance variability, unpredictable locations, and frequent small-sized lesions of interest~\cite{Litjens2017survey,yan_2018_deeplesion}. In particular, two key aspects requiring further research are (1) the best means to effectively process the 3D volumetric data (since small and critical tumors require 3D imaging context to be differentiated) and (2) to more accurately regress the tumor's 3D bounding box. This work makes significant contributions towards both aims. 
   
\Ac{CT} scans are volumetric, so incorporating 3D context is the key in recognizing lesions. As a direct solution, 3D \acp{CNN} have achieved good performance for lung nodule detection~\cite{DBLP:conf/miccai/DingLHW17,DBLP:journals/tbe/DouCYQH17}. However, due to GPU memory constraints, shallower networks and smaller input dimensions are used~\cite{DBLP:conf/miccai/DingLHW17,DBLP:journals/tbe/DouCYQH17}, which may limit the performance for more complicated detection problems. For instance, \ac{ULD}~\cite{miccai/ShaoGMLZ19,isbi/TangYTLXS19,miccai/YanTPSBLS19,miccai/ZlochaDG19}, which aims to detect many lesions types with diverse appearances from the whole body, demands wider and deeper networks to extract more comprehensive image features. To resolve this issue, 2.5D networks have been designed~\cite{LesionHarvesterJZCai,miccai/ShaoGMLZ19,isbi/TangYTLXS19,DBLP:conf/miccai/YanBS18,miccai/YanTPSBLS19,miccai/ZlochaDG19} that use deep 2D \acp{CNN} with ImageNet pre-trained weights and fuse image features of multiple consecutive axial slices. Nevertheless, these methods do not fully exploit 3D information since their 3D related operations operate sparsely at only selected network layers via convolutional-layer inner products. 2.5D models are also inefficient because they process CT volumes in a slice-by-slice manner. Partially inspired by~\cite{cvpr/Carreira17I3D,iccv/QiuYM17,corr/abs-1911-10477}, we propose applying \ac{P3DC} backbones to efficiently process 3D images. This allows our \ac{VULD} framework to fully exploit 3D context while re-purposing off-the-shelf deep 2D network structures and inheriting their large capacities to cope with lesion variances.

Good lesion detection performance also relies on accurate bounding box regression. But, some lesions, \eg{} liver lesions, frequently present vague boundaries that are hard to distinguish from background. Most existing anchor-based~\cite{renNIPS15fasterrcnn} and anchor-free~\cite{DBLP:journals/corr/abs-1904-07850,iccv/Tian19FCOS} algorithms rely on features extracted from the proposal \emph{center} to predict the lesion's extent. This is sub-optimal since lesion \emph{boundary} features should intuitively be crucial for this task. To this end, we adopt and enhance the RepPoint algorithm~\cite{iccv/YangLHWL19}, which generates a point set to estimate bounding boxes, with each point fixating on a representative part. Such a point set can drive more finely-tuned bounding box regression than traditional strategies, which is crucial for accurately localizing small lesions. Different from  RepPoint, we propose \ac{SPR}, which uses a novel triplet-base appearance regularization to force regressed points to move towards lesion boundaries. This allows for an even more accurate regression. 

In this work, we advance both volumetric detection and bounding box regression using deep volumetric \acp{P3DC} and effective \ac{SPR}, respectively. We demonstrate that our \ac{P3DC} backbone can outperform \acl{SOTA} 2.5D and 3D detectors on the public large-scale NIH DeepLesion dataset~\cite{yan_2018_deeplesion}, \eg{} we increase the strongest baseline's sensitivity of detecting small lesions from $22.4\%$ to $30.3\%$ at 1 \ac{FP} per \ac{CT} volume. When incorporating \ac{SPR}, our \ac{VULD} outperforms the best baseline~\cite{LesionHarvesterJZCai} by $>4\%$ sensitivity for all operating points on \ac{FROC}. We also evaluate \ac{VULD} on an extremely challenging dataset (574 patient studies) of \ac{HCC} liver lesions collected from archives in Chang Cung Memorial Hospital. Many patients suffer from cirrhosis, which make \ac{HCC} detection extremely difficult. \ac{P3DC}  alone accounts for $63.6\%$ sensitivity at 1 \ac{FP} per \ac{CT} volume. Adding \ac{SPR} boosts this sensitivity to $69.2\%$. Importantly, for both the DeepLesion and in-house \ac{HCC} dataset, our complete \ac{VULD} framework provides the largest performance gains for small lesions, which are the easiest to miss by human readers and thus should be the focus for any detection system.

\section{Method}

\begin{figure}[t!]
\centering
\includegraphics[width=\textwidth, trim=0.1in 0.1in 0.1in 0.1in, clip]{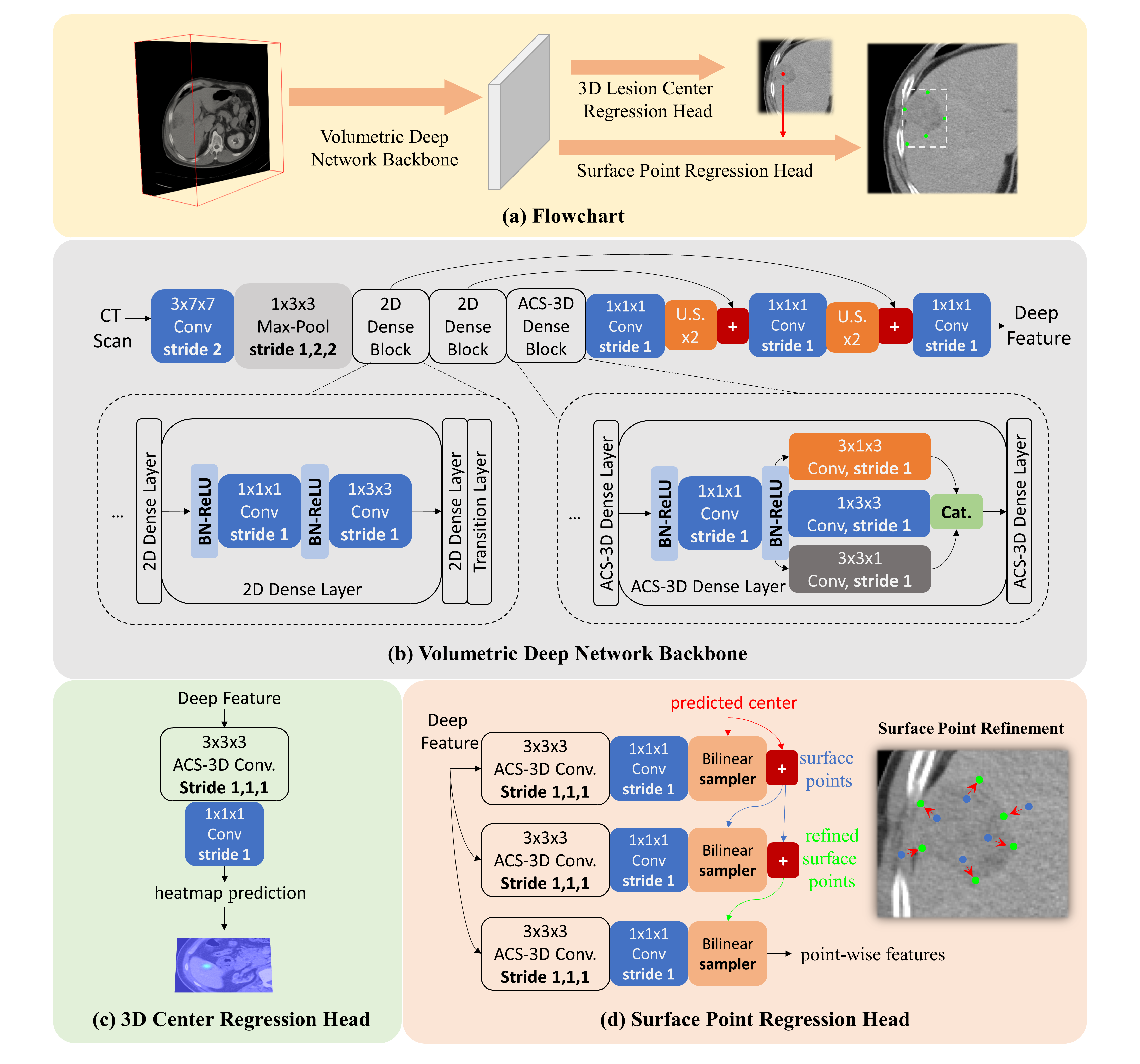}
\caption{Overview of \ac{VULD}. We show (a) the complete workflow; (b) the detailed \acf{P3DC} backbone; (c) 3D lesion center regression head; and (d) \acf{SPR} head for bounding box generation.} %Best viewed in color.}
\label{fig:framework}
\end{figure}

\ac{VULD} follows a one-stage anchor-free detection workflow~\cite{LesionHarvesterJZCai,DBLP:journals/corr/abs-1904-07850}, which is simple but has yielded state-of-the-art performance on DeepLesion~\cite{LesionHarvesterJZCai}. As shown in \Fig{framework}, \ac{VULD} takes volumetric \ac{CT} scans as inputs and extracts deep convolutional features with its \ac{P3DC} backbone. The extracted features are then fed into \ac{VULD}'s 3D center regression and \ac{SPR} heads to generate center coordinates and  surface points, respectively. 

\subsection{\acs{P3DC} Backbone}

\ac{VULD} relies on a deep volumetric \ac{P3DC} backbone. To do this, we build off of DenseNet-121~\cite{DBLP:conf/cvpr/HuangLMW17}. Specifically, we first remove the fourth dense block as we found this truncated version performs better with DeepLesion. The core strategy of \ac{VULD} is to keep front-end processing to 2D, while only converting the third dense block of the truncated DenseNet-121 to 3D using \acp{P3DC}. This strategy is consistent with \cite{miccai/YanTPSBLS19}, which found that introducing 3D information at higher layers is preferred to lower layers. Using $N$ to denote convolutional kernel sizes throughout, for the first two dense blocks the weight parameters,  $(c_{o},c_{i},N,N)$, are reshaped to $(c_{o},c_{i},1,N,N)$ to process volumetric data slice-by-slice. When processing dynamic \acp{CT} with multiple contrast phases, \eg{}, our in-house dataset, we stack the multi-phase input and inflate the weight of the first convolutional kernel along its second dimension~\cite{cvpr/Carreira17I3D}. 

To implement 3D processing, we convert the third dense block and task-specific heads and investigate several different options for \acp{P3DC}, which include \ac{I3D}~\cite{cvpr/Carreira17I3D}, \ac{ST3D}~\cite{iccv/QiuYM17}, and \ac{ACS3D}~\cite{corr/abs-1911-10477}.
\begin{figure}[t!]
\centering
\includegraphics[width=1.0\textwidth]{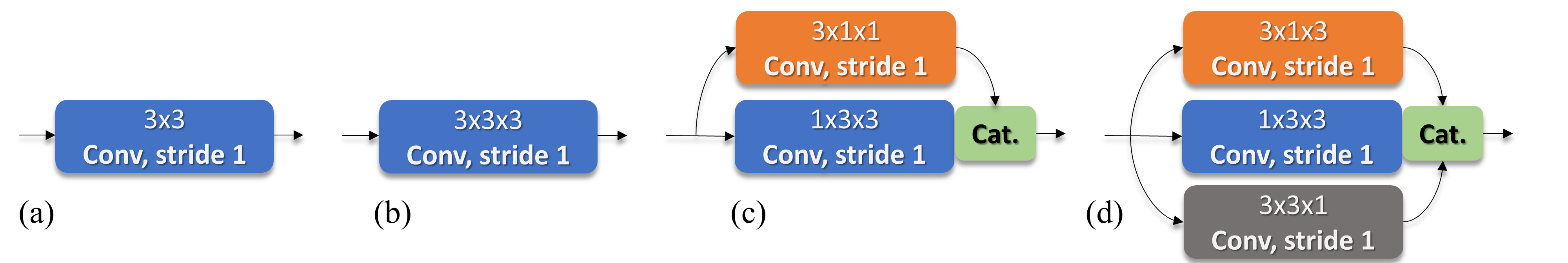}
\caption{Options to transfer the 2D convolutional layer (a) to volumetric 3D convolutions: (b) \acl{I3D}~\cite{cvpr/Carreira17I3D}, (c) \acl{ST3D}~\cite{iccv/QiuYM17}, and (d) \acl{ACS3D}~\cite{corr/abs-1911-10477}.} \label{fig:psuedo3d} 
\end{figure}
These options are depicted in \Fig{psuedo3d}. \ac{I3D}~\cite{cvpr/Carreira17I3D} simply duplicates 2D kernels along the axial (3D) direction and downscales weight values by the number of duplications. Thus, \ac{I3D} produces true 3D kernels. \ac{ST3D}~\cite{iccv/QiuYM17} first reshapes $(c_o,c_i,N,N)$ kernels into $(c_o,c_i,1,N,N)$ to act as ``spatial'' kernels and introduces an extra $(c_o,c_i,N,1,1)$ kernel as the ``temporal'' kernel. The resulting features from both are fused using channel-wise concatenation. There are alternative \ac{ST3D} configurations; however, the parallel structure of \Fig{psuedo3d}(c) was shown to be best in a liver segmentation study~\cite{ijcai/ZhangXZCXS19}. \ac{ACS3D}~\cite{corr/abs-1911-10477} splits the kernel $(c_o,c_i,N,N)$ into axial $(c_{oa},c_i,N,N)$, coronal $(c_{oc},c_i,N,N)$, and sagittal $(c_{os},c_i,N,N)$ kernels, where $c_o=c_{oa}+c_{os}+c_{oc}$. Thereafter, it reshapes the view-specific kernels correspondingly into $(c_{oa},c_i,1,N,N)$, $(c_{oc},c_i,N,1,N)$, and $(c_{os},c_i,N,N,1)$. Like \ac{ST3D}, \ac{ACS3D} fuses the resulting features using channel-wise concatenation. Compared to the extra temporal-kernels introduced by \ac{ST3D}, \ac{ACS3D} requires no extra model parameters, keeping the converted model light-weight. In our implementation, we empirically set the ratio of $c_{oa}:c_{oc}:c_{os}$ to $8:1:1$ as the axial plane usually holds the highest resolution.

\Ac{VULD} has two task-specific network heads, one to locate the lesion centers and one to regress surface points. Before inputting the deep volumetric features into the heads, we use an \ac{FPN}~\cite{DBLP:conf/cvpr/LinDGHHB17} with three $(c_o,c_i,1,1,1)$ convolutional layers to fuse outputs of the dense blocks, which helps \ac{VULD} to be robust to lesions with different sizes. Focusing first on the center regression head, it takes the output of the \ac{FPN} (\ie{} ``deep feature'' in \Fig{framework}) and processes it with an \ac{ACS3D} convolutional layer followed by a $(1,c_i,1,1,1)$ convolutional layer. Both layers are randomly initialized. Like CenterNet~\cite{DBLP:journals/corr/abs-1904-07850}, the output is a 3D heat map, $\hat{Y}$, that predicts lesion centers. Ground-truth heat map, $Y$, is generated as a Gaussian heat map with the radius in each dimension set to half of the target lesion's width, height, and depth. We use focal loss~\cite{LesionHarvesterJZCai,DBLP:conf/iccv/LinGGHD17,DBLP:journals/corr/abs-1904-07850} to train the center regression head:
\begin{equation} \label{eq:centerLoss}
    \mathcal{L}_{ctr} = \frac{-1}{m} \sum_{xyz}
    \begin{cases}
        (1 - \hat{Y}_{xyz})^{\alpha} 
        \log(\hat{Y}_{xyz}) & \!\text{if}\ Y_{xyz}=1 \\
        \begin{array}{c}
        (1-Y_{xyz})^{\beta} 
        (\hat{Y}_{xyz})^{\alpha}   
        \log(1-\hat{Y}_{xyz})
        \end{array}
        & \!\text{otherwise}
    \end{cases} \textrm{,}
\end{equation}
where $m$ is the number of lesions in the \ac{CT} and $\alpha=2$ and $\beta=4$ are focal-loss hyper-parameters~\cite{DBLP:journals/corr/abs-1904-07850}. The ground-truth heat map is $<1$ everywhere except at the lesion center voxel. Like recent work~\cite{LesionHarvesterJZCai}, when possible we also exploit hard negatives by generating negative-valued heatmaps in $Y$, which will magnify their loss contributions more than $0$-valued regions. See Cai \etal{}~\cite{LesionHarvesterJZCai} for more details.

\subsection{Surface Point Regression}

The \ac{P3DC} backbone and center regression head are effective at locating lesions. But, once the lesion is located its extent must also be determined. To do this, we directly regress a 3D point set (actually offsets from the center point), using backbone features located at the center point:
\begin{equation}
\mathcal{P} = \{(x_k, y_k, z_k)\}_{k=1}^{n}, 
\end{equation}
where $n$ is the total number of points. This requires a $1\times 1 \times 1$ convolution with $3n$ outputs. Empirically, we find $n=16$ delivers the best results. Because $\mathcal{P}$ is computed from center-point features, it may suffer from inaccuracies. Thus, we also compute offsets to refine $\mathcal{P}$: 
\begin{equation}
\mathcal{P}_r = \{(x_k + \Delta x_k, y_k + \Delta y_k, z_k + \Delta z_k)\}_{k=1}^{n}, \label{eqn:offset}
\end{equation}
where $\{(\Delta x_k, \Delta y_k, \Delta z_k)\}$ are the predicted offsets of the refined surface points. To do this, for each location in $\mathcal{P}$, we bilinearly interpolate corresponding backbone features and regress location-specific offsets. This only requires a $1\times 1 \times 1$ convolution with $3$ outputs. To actually supervise the $\mathcal{P}$ and $\mathcal{P}_{r}$ regression, we compute their minimum and maximum coordinates and ensure they match with the ground-truth bounding box. More formally, if we denote the ground-truth box using its top-right-front and bottom-left-rear corners $\{(x_{trf}, y_{trf}, z_{trf}),$ $ (x_{blr}, y_{blr}, z_{blr})\}$, the regression of $\mathcal{P}$ and $\mathcal{P}_{r}$ can be trained using the following loss: 
\begin{multline}
\mathcal{L}_{pts} = \sum_{i \in (x, y, z)} 
|i_{blr} - \min_{1\leq k \leq n}(i_k)| + |i_{trf} - \max_{1\leq k \leq n}(i_k)| + \\
|i_{blr} - \min_{1\leq k \leq n}(i_k + \Delta i_k)| + |i_{trf} - \max_{1\leq k \leq n}(i_k + \Delta i_k)| \mathrm{.} \label{eqn:pr}
\end{multline}

One important limitation of \eqref{eqn:pr} is that ellipsoid lesions do not fit perfectly in cuboid boxes. As a result, regressed points may still satisfy \eqref{eqn:pr} if they lay outside the lesion, but still inside the box. Such points may be more prone to produce inaccurate offsets, \ie{} \eqref{eqn:offset}, during inference. To address this, we propose an appearance-based similarity constraint to encourage points to only fixate on lesion surfaces so that the point set can represent fine-grained lesion geometry correctly. The idea is to force surface-point appearance to be more similar to regions inside the lesion than to those outside it. This constraint is achieved by adding a triplet-loss with the lesion center as the positive anchor (inside) and box corners as negative anchors (outside). Specifically, we compute \emph{point-wise} features from the center and eight corners of the bounding box with bilinear sampling and denote them as $a^p$ and $\{a^n_{j}\}_{j=1}^8$, respectively. We also extract point-wise features from $P_r$: $\{a_{k}\}_{k=1}^n$. The triplet-loss is then formulated as 
\begin{equation}
  \mathcal{L}_{tri} = \frac{1}{m} \sum_{k=1}^{n}\sum_{j=1}^{8} \max(0, \|a^p - a_k\|_2 - \|a^p - a^n_j\|_2 + 1).
\end{equation}
With the supervision of $\mathcal{L}_{pts}$ and $\mathcal{L}_{tri}$, we expect surface points will either move toward lesion surfaces or to the center. This constitutes our \acf{SPR}. The extracted point-wise features are designed to be semantic in nature (healthy versus lesion tissue). Thus, complex lesion appearances, \eg{}, cavitations, should be mapped to a similar semantic space. We optimize the \ac{SPR} together with the center regression head by minimizing a joint loss function :
\begin{equation}
  \mathcal{L} = \mathcal{L}_{ctr} + 0.1(\mathcal{L}_{pts} + \mathcal{L}_{tri}).
\end{equation}

\subsection{Implementation Details}
We implement our system in Pytorch~\cite{nips/PaszkeGMLBCKLGA19} on four {NVIDIA} Quadro {RTX} 6000 {GPUs}. The \ac{P3DC} backbone weights were initialized with the pre-trained Lesion Harvester weights~\cite{LesionHarvesterJZCai}, which were trained using the official DeepLesion data split so there is no data leakage. We also tried ImageNet-pretrained weights and random initialization, but performance was not as good. All other layers were randomly initialized.  The \ac{FPN}'s output, \ie{}, ``deep feature'' in \Fig{framework}, has 512 channels. In the task-specific heads, each \ac{ACS3D} layer consists of an \ac{ACS3D} convolutional layer with a kernel size of 3 and $c_{ao}+c_{co}+c_{so}=256$ . The output channels of the lesion center heat map, $\mathcal{P}$, $\mathcal{P}_{r}$, and point-wise features are 1, 48 (16 points), 3, and 128, respectively. We adopt the Adam~\cite{corr/KingmaB14} optimizer and set a base learning rate to 0.0001, which was reduced by a factor of 10 after the validation loss reached its minimum value.

\section{Experimental Results}

\noindent \textbf{Datasets.} We evaluate our approach on two datasets. \textbf{DeepLesion}~\cite{cvpr/yan18deeplesion} is a large-scale benchmark for \ac{ULD} that comprises 32,735 retrospectively clinically annotated lesions from 10,594 CT scans of 4,427 unique patients. Many works report performance on DeepLesion, but most are either 2D~\cite{isbi/TangYTLXS19,miccai/ZlochaDG19,miccai/ShaoGMLZ19} or 2.5D~\cite{DBLP:conf/miccai/YanBS18,miccai/YanTPSBLS19}. We use the 3D annotations and hard-negatives from~\cite{LesionHarvesterJZCai} to both train and evaluate DeepLesion. The volumetric test set of DeepLesion~\cite{LesionHarvesterJZCai} includes 272 fully-annotated sub-volumes and more accurately reflects the 3D lesion detection performance. \textbf{\acs{HCC} Liver Dataset:} We also evaluate on our in-house dataset of $574$ dynamic \ac{CT} studies of patients with \ac{HCC} liver lesions. \Ac{HCC} is one of the most fatal cancers and detection at early stages is crucial. However, \ac{HCC} often co-occurs with liver fibrosis, challenging lesion discovery. Human sensitivities have been reported to be $48\sim57\%$ for small-sized lesions~\cite{Addley_2011}. We randomly split the dataset patient-wise into $384$, $92$, and $98$ studies for training, validation, and testing, respectively.  

\noindent \textbf{Evaluation and Comparison Methods.}
A detected bounding-box is regarded as correct when the 3D-\acs{IoU} between the detected box and a ground-truth box exceeds $0.3$. The \ac{FROC} is used for evaluation. We first evaluate different \ac{P3DC} backbones: \ac{ST3D}, \ac{I3D}, and \ac{ACS3D}. We also test a shallow fully-3D UNet~\cite{miccai/CicekALBR16} backbone within the CenterNet~\cite{DBLP:journals/corr/abs-1904-07850} framework and also against the 2.5D Lesion Harvester~\cite{LesionHarvesterJZCai}, which reports the highest performance to date for the DeepLesion dataset. These two competitors directly regress a lesion's size using features sampled from the predicted lesion center and can also naturally learn from hard-negatives~\cite{LesionHarvesterJZCai}. In addition, we also report results using CenterNet (2D)~\cite{DBLP:journals/corr/abs-1904-07850}, Faster R-CNN (2.5D)~\cite{renNIPS15fasterrcnn}, and MULAN (2.5D)~\cite{miccai/YanTPSBLS19}, drawn from Cai \etal{}'s experiments~\cite{LesionHarvesterJZCai}. This represents a comprehensive comparison across many different detector variants. To measure the impact of our proposed \ac{SPR}, we also implement \ac{VULD} with \acf{PR}~\cite{corr/abs-1912-11473}  that foregoes the appearance-based triplet loss. Finally, we evaluate our proposed \ac{VULD} framework: \ac{P3DC} + \ac{SPR}. 

\begin{table}[t!]
\scriptsize
\centering 
\caption{Sensitivities (\%) at various \acp{FP} per \ac{CT} volume.}  
\label{tbl:froc}
\begin{tabular}{|l|c|c|c|c|c|c|c|c|c|c|}
\hline
\multirow{2}{*}{\textbf{Method}} & \multirow{2}{*}{\textbf{backbone}} & \multicolumn{8}{c|}{\textbf{FPs per Volume}} & \multirow{2}{*}{\textbf{Avg.}} \\
\cline{3-10}
& & 0.25 & 0.50 & 0.75 & 1.00 & 1.25 & 1.50 & 1.75 & 2.00 &  \\
\hline 
\multicolumn{11}{|c|}{\textit{DeepLesion volumetric test set}} \\
\hline 
CenterNet-3D & 3D UNet                                               & 9.6 & 14.1 & 16.7 & 18.9 & 20.3 & 22.2 & 23.5 & 24.9 & 18.7 \\
Faster {R-CNN}~\cite{renNIPS15fasterrcnn} & 2.5D DenseNet-121        & 9.0 & 14.8 & 19.8 & 25.6 & 29.3 & 32.8 & 35.5 & 36.7 & 25.4 \\
CenterNet~\cite{DBLP:journals/corr/abs-1904-07850} & 2D DenseNet-121 & 15.0 & 19.8 & 24.3 & 28.5 & 31.2 & 33.3 & 35.0 & 36.6 & 27.9\\  
MULAN~\cite{miccai/YanTPSBLS19} & 2.5D DenseNet-121                  & 14.5 & 20.8 & 25.6 & 31.0 & 34.4 & 38.1 & 40.3 & 42.8 & 30.9 \\
Lesion Harvester~\cite{LesionHarvesterJZCai} & 2.5D DenseNet-121     & 15.8 & 24.6 & 28.4 & 32.7 & 35.5 & 37.5 & 39.8 & 41.0 & 31.9 \\
\acs{P3DC} & \acs{I3D}             & \textbf{22.3} & 27.7 & 32.7 & 36.5 & 38.3 & 39.5 & 41.4 & 43.0 & 35.1 \\
\acs{P3DC} & \acs{ST3D}            & 18.8          & 26.7 & 30.5 & 32.7 & 35.5 & 37.3 & 39.1 & 41.2 & 32.7 \\
\acs{P3DC} & \acs{ACS3D}           & 19.8          & 27.5 & 32.2 & 35.5 & 38.9 & 41.1 & 41.8 & 43.1 & 34.9 \\
\acs{P3DC} & \acs{ACS3D}+\acs{PR}  & 20.4          & 26.3 & 31.0 & 34.4 & 37.4 & 40.0 & 41.3 & 42.1 & 34.1 \\
\acs{P3DC} & \acs{ACS3D}+\acs{SPR} & 20.1 & \textbf{29.1} & \textbf{34.4} & \textbf{37.1} & \textbf{40.3} & \textbf{42.1} & \textbf{43.6} & \textbf{45.1} & \textbf{36.4} \\
\hline  
\multicolumn{11}{|c|}{\textit{\ac{HCC} Liver test set}} \\
\hline  
\acs{P3DC} & \acs{ACS3D}           & 50.5          & 57.0          & 61.7 & 63.6          & 67.3 & 71.0 & 71.0 & 71.0 & 64.1 \\
\acs{P3DC} & \acs{ACS3D}+\acs{PR}  & \textbf{57.9} & \textbf{65.4} & 68.2 & \textbf{69.2} & 70.1 & 71.0 & 72.9 & 73.8 & 68.5 \\
\acs{P3DC} & \acs{ACS3D}+\acs{SPR} & 55.1 & 64.5 & \textbf{69.2} & \textbf{69.2} & \textbf{72.0} & \textbf{76.6} & \textbf{77.6} & \textbf{77.6} & \textbf{70.2} \\
\hline  
\end{tabular}
\end{table}

\noindent \textbf{Results.} 
In \Table{froc}, we compare our proposed approach against alternative approaches. Using FROC analysis, the average sensitivities on DeepLesion are: CenterNet-2D $27.9\%$; CenterNet-3D $18.7\%$; Faster R-CNN $25.4\%$; MULAN $27.9\%$; Lesion Harvester $31.9\%$, and our strongest \acs{P3DC} variant $36.4\%$. As can been seen, \acs{P3DC} significantly outperforms the previous \ac{SOTA} Lesion Harvester and MULAN methods by $4.5\%$ and $8.5\%$, respectively, which validates the effectiveness of \ac{P3DC} over its 2.5D counterparts.  

From \Table{froc}, we also observe that adding the original \ac{PR} method actually underperforms the baseline \acs{P3DC}. This in fact motivated our development of \ac{SPR}. The \ac{PR} method lacks explicit constraints on point locations, making it challenging to automatically learn effective point-wise feature from CT images. In contrast, \ac{SPR} introduces surface constraints to force the regressed points to distribute onto lesion surfaces. Tests on our in-house dataset also confirms that our proposed \ac{SPR} can improve sensitivities on \ac{HCC} liver lesion detection.

While these results demonstrate the value of our \ac{P3DC} backbone and \ac{SPR} bounding-box regression, even more convincing conclusions can be drawn when analyzing performance based on lesion size. In DeepLesion, we use 2$cm$ and 5$cm$ as cut-off sizes. However, our \ac{HCC} liver dataset has hardly any lesions smaller than 2$cm$, so we only stratify based on a 5$cm$ cut-off. As \Table{size} indicates, compared to Lesion Harvester, our \ac{P3DC} backbone can yield improvements of $~7\%$ sensitivity for small-size lesions in DeepLesion. These are the most critical lesions to detect, since these are the easiest for human observers to miss. Adding the \ac{SPR} boosts small-size performance even further, indicating that \ac{SPR}'s aggregation of boundary features can produce improved fine-grained bounding boxes. Moving to the \ac{HCC} dataset, our \ac{SPR} can produce boosts in sensitivity of over $4\%$ compared to direct CenterNet-style regression, further validating our \ac{SPR} regression strategy. These are clinically significant performance improvements. Visual examples can be found in \Fig{visualization}, and our supplementary material, depicting the process of \ac{SPR}'s more refined regression of bounding box extents.  

\begin{table}[t!]
\scriptsize
\centering 
\caption{Size-stratified sensitivities (\%) at FP$=1$ per \ac{CT} volume.$^{\dagger}$: \ac{P3DC}+\acs{PR} produces \acp{FP} with high confidences, thus at \ac{FP}$=1$, it has lower sensitivity than \ac{P3DC}+\acs{SPR} on \ac{HCC} Liver.}
\label{tbl:size}
\begin{tabular}{l||cccc||ccc}
\hline
& \multicolumn{4}{c||}{\textbf{DeepLesion}} & \multicolumn{3}{c}{\textbf{\acs{HCC} Liver}} \\
Lesion size ($cm$) & $<$2 & 2$\sim$5 & $>$5 & All & $<=$5 & $>$5 & All \\
%                  & 541 & 255 & 78 & 874 \\
Distribution (\%)  & 62\% & 29\% & 9\% & 100\% & 65\% & 35\% & 100\% \\
\hline 
CenterNet-3D & 16.5 & 38.8 & 30.8 & 18.9 & & & \\
Lesion Harvester~\cite{LesionHarvesterJZCai}    &  22.4 & \textbf{67.1} & \textbf{75.6} & 32.7 & & & \\
\ac{P3DC} (\ac{ACS3D})                            &  29.9 & 62.4 & 57.7 & 35.5 & 61.4 & 86.5 & 63.6 \\
\ac{P3DC} (\ac{ACS3D}) + \acs{PR}                 &  28.7 & 64.7 & 60.3 & 34.4 & 58.6$^{\dagger}$ & \textbf{94.6} & 69.2 \\ 
\ac{P3DC} (\ac{ACS3D}) + \acs{SPR}             ~~~&  \textbf{30.3} & 63.9 & 62.8 & \textbf{37.1} & \textbf{65.7} & \textbf{94.6} & \textbf{69.2} \\
\hline
\end{tabular}
\end{table}

\begin{figure}[t!]
\centering 
\includegraphics[width=0.98\textwidth]{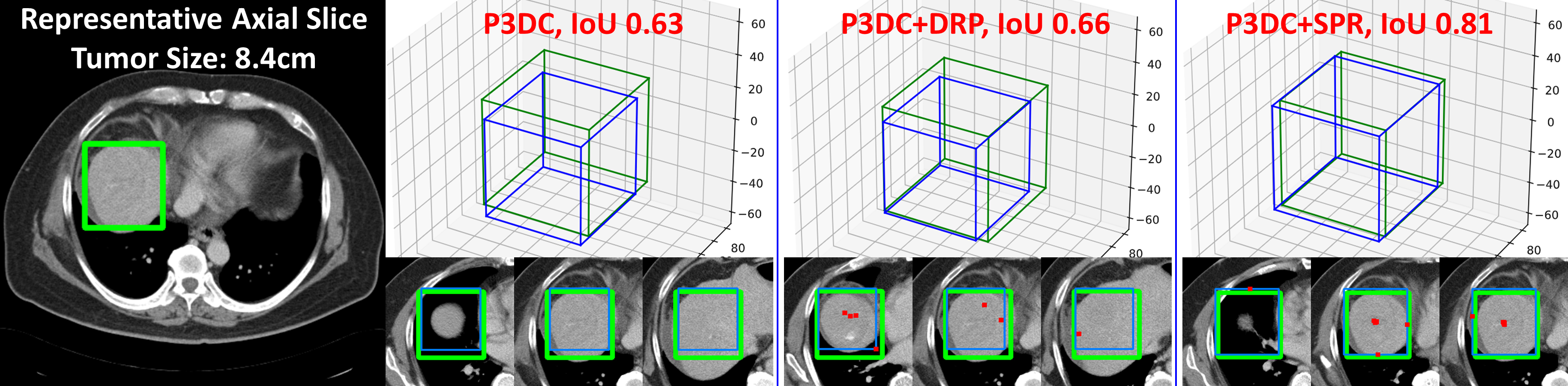}
\caption{Visualization of different methods. We show an instance of liver tumor overlaid with its ground-truth box in the 1$^{st}$ column. In the 2$^{nd}$, 3$^{rd}$, and 4$^{th}$ columns, we show the detection results from \ac{P3DC} with general box regression, \ac{P3DC}$+$\ac{PR}, and \ac{P3DC}$+$\ac{SPR}, respectively. For each example, we display the result in 3D and show three representative axial slices. We render the ground-truth box in {\bf Green}, the detection results in {\bf Blue}, and the regressed surface points, when applicable, in {\bf Red}. Best viewed in color.}
\label{fig:visualization}
\end{figure}

\section{Conclusion} 

In this work, we tackle challenges of lesion detection in \ac{CT} scans by proposing a very deep volumetric lesion detection model \ac{VULD}. It processes \ac{CT} scans directly in 3D so as to fully incorporate 3D context for better performance. It has very deep backbones with large capacities so that it can handle lesions with large appearance variability. Its \acl{SPR} head can effectively estimate the 3D lesion spatial extents. It also generalize well with small-scaled medical datasets as it is light-weight and can be initialized with pre-trained 2D networks. Compared with 2D, 2.5D, and fully 3D variants, our method is superior in accuracy, model size, and speed (see our supplementary material). The proposed \ac{VULD} acheived new \ac{SOTA} performance on the large-scale NIH DeepLesion dataset. It has also validated its generalization capability on our in-house \ac{HCC} liver dataset. 

\bibliographystyle{splncs04}
\bibliography{paii}

\newpage
\section*{Supplementary Material}
%--- Add this into supplementary  
\subsection*{Computational Efficiency} 
We compared \ac{P3DC} with 2D, 2.5D, and 3D CNNs, all using a DenseNet-121 backbone. The numbers of parameters are: \ac{P3DC} ($12.7M$), 2D ($8.9M$), 2.5D ($9.7M$), and 3D ($18.7M$). Using an input volume size of 32$\times$256$\times$256, the FLOPs are \ac{P3DC} ($945$ GFLOPs), 2D ($1280$ GFLOPs), 2.5D ($1989$ GFLOPs), and 3D ($1626$ GFLOPs). \ac{P3DC} is more efficient than 2D and 2.5D methods because the latter two predict a 3D volume slice-by-slice, while \ac{P3DC} can infer a sub-volume at a time. In summary, \ac{P3DC} is superior in accuracy, model size, and speed. %We will include a table of these numbers in the supplementary material and mention the main results in the body text  

\begin{figure}[h!]
\centering 
\includegraphics[width=0.90\textwidth]{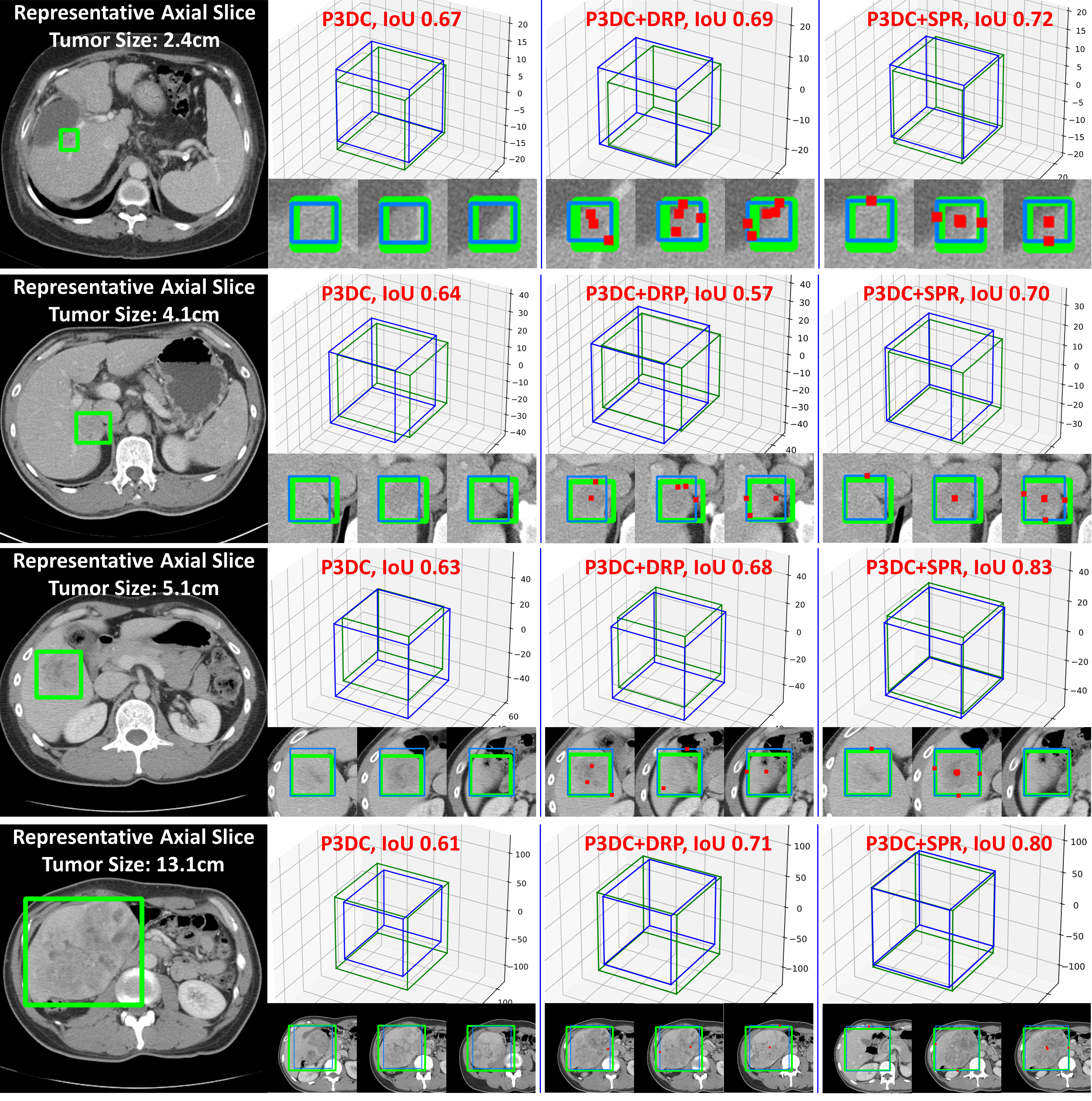}
\caption{More visualization examples for 3D liver tumor detection from our in-house \acs{HCC} liver dataset. We show instances of liver tumors overlaid with their ground-truth boxes in the 1$^{st}$ column. In the 2$^{nd}$, 3$^{rd}$, and 4$^{th}$ columns, we show the detection results from \ac{P3DC} with direct bounding-box regression, \ac{P3DC}$+$\ac{PR}, and our proposed \ac{P3DC}$+$\ac{SPR}, respectively. For each example, we display the result in 3D and show three representative axial slices. We render the ground-truth boxes in {\bf Green}, the detection results in {\bf Blue}, and the regressed surface points, when applicable, in {\bf Red}. Best viewed in color.}
\label{fig:visualization}
\end{figure}

\begin{figure}[h!]
\centering 
\includegraphics[width=0.90\textwidth]{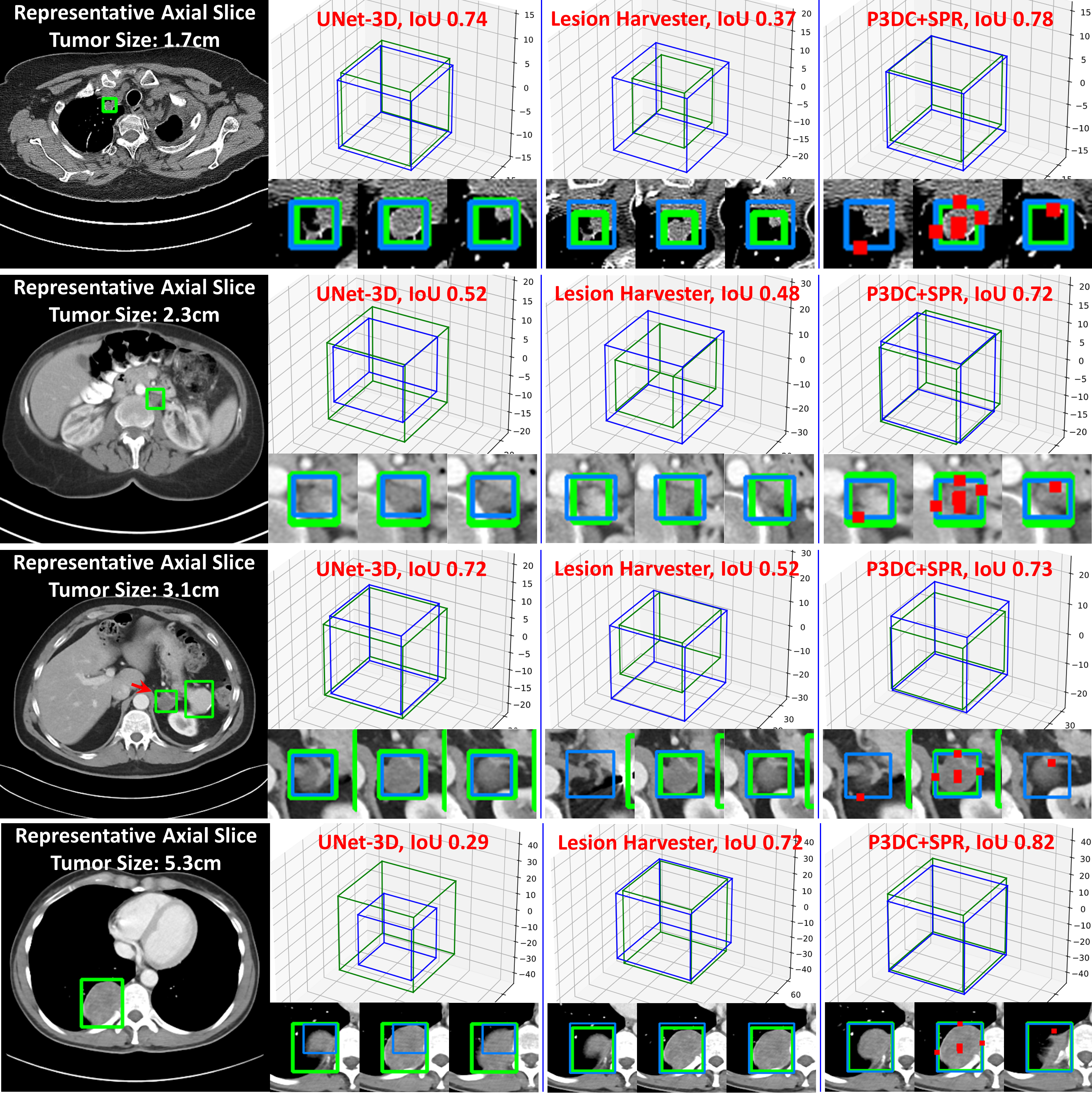}
\caption{More visualization examples for 3D lesion detection from the DeepLesion dataset. We show instances of lesions overlaid with their ground-truth boxes in the 1$^{st}$ column. In the 2$^{nd}$, 3$^{rd}$, and 4$^{th}$ columns, we show the detection results from UNet-3D, Lesion Harvester, and our proposed \ac{P3DC}$+$\ac{SPR}, respectively. For each example, we display the result in 3D and show three representative axial slices. We render the ground-truth boxes in {\bf Green}, the detection results in {\bf Blue}, and the regressed surface points, when applicable, in {\bf Red}. Best viewed in color.}
\label{fig:visualization}
\end{figure}

\end{document}